\documentclass{article}

\usepackage[preprint]{neurips_2022}

\usepackage[utf8]{inputenc} 
\usepackage[T1]{fontenc}    
\usepackage{hyperref}       
\usepackage{url}            
\usepackage{booktabs}       
\usepackage{amsfonts}       
\usepackage{nicefrac}       
\usepackage{microtype}      
\usepackage{xcolor}         
\usepackage{soul}
\usepackage{tablefootnote}
\usepackage[normalem]{ulem}
\usepackage{multirow}
\usepackage{makecell}
\usepackage{enumitem}
\usepackage{adjustbox}
\usepackage{amsmath}
\usepackage{array}
\usepackage{graphicx}
\usepackage{comment}
\usepackage{booktabs}
\usepackage{xcolor,colortbl} 
\usepackage{algorithm}

\usepackage{bbm}
\usepackage{tabularx}
\usepackage{hyperref}

\usepackage{graphicx}
\usepackage{subcaption}
\usepackage{arydshln}
\usepackage{array}
\usepackage{booktabs}
\usepackage{pifont}
\usepackage{cleveref}
\crefname{section}{§}{§§}
\usepackage{amsthm}
\usepackage{todonotes}
\usepackage{color}
\usepackage{amsmath, bm, amsfonts}
\usepackage{newfloat}
\usepackage{listings}
\usepackage{multirow}

\newcommand{\benchmarkname}{{\usefont{T1}{ppl}{m}{n}AGIEval}}

\title{AGIEval: A Human-Centric Benchmark for Evaluating Foundation Models}

\author{
    Wanjun Zhong\thanks{indicates equal contribution. Yaobo Liang and Nan Duan are the corresponding authors.},~~Ruixiang Cui$^{*}$,~~Yiduo Guo,~~Yaobo Liang,~~Shuai Lu,~~Yanlin Wang\\\textbf{~~Amin Saied,~~Weizhu Chen and Nan Duan}\\
    Microsoft
    \\
    \texttt{\{t-wzhong, v-ruicui, v-yiduoguo, yaobo.liang, shuailu\}@microsoft.com} \\ 
    \texttt{\{yanlwang, Amin.Saied, wzchen, nanduan\}@microsoft.com}\\
}

\begin{document}

\maketitle

\begin{abstract}
Evaluating the general abilities of foundation models to tackle human-level tasks is a vital aspect of their development and application in the pursuit of Artificial General Intelligence (AGI). Traditional benchmarks, which rely on artificial datasets, may not accurately represent human-level capabilities. In this paper, we introduce \benchmarkname, a novel benchmark specifically designed to assess foundation model in the context of human-centric standardized exams, such as college entrance exams, law school admission tests, math competitions, and lawyer qualification tests. 
We evaluate several state-of-the-art foundation models, including GPT-4, ChatGPT, and Text-Davinci-003, using this benchmark. 
Impressively, GPT-4 surpasses average human performance on SAT, LSAT, and math competitions, attaining a 95\% accuracy rate on the SAT Math test and a 92.5\% accuracy on the English test of the Chinese national college entrance exam. This demonstrates the extraordinary performance of contemporary foundation models.  
In contrast, we also find that GPT-4 is less proficient in tasks that require complex reasoning or specific domain knowledge. Our comprehensive analyses of model capabilities (understanding, knowledge, reasoning, and calculation) reveal these models' strengths and limitations, providing valuable insights into future directions for enhancing their general capabilities.
By concentrating on tasks pertinent to human cognition and decision-making, our benchmark delivers a more meaningful and robust evaluation of foundation models' performance in real-world scenarios\footnote{The data, code, and all model outputs are released in \url{https://github.com/ruixiangcui/AGIEval}}.

\end{abstract}
\section{Introduction}
In recent years, large foundation models, such as the latest large language models (LLMs) ChatGPT\footnote{\url{https://chat.openai.com/chat}} and GPT-4~\citep{openai2023gpt4}, have exhibited remarkable versatility and adaptability, with plethora of applications spanning various domains as a decision-making assistant, from processing daily events to assisting in specialized fields such as law, medicine, and finance. 
With these advancements, AI systems are inching closer to achieving Artificial General Intelligence (AGI). 
As these AI systems continue to evolve and become more integrated into our daily lives, it is essential to effectively assess their general abilities in handling human-centric tasks, identify potential shortcomings, and ensure that they can handle complex, human-centric tasks effectively. Moreover, evaluating their reasoning abilities is also crucial to ensure their reliability and trustworthiness across diverse settings.

Traditional benchmarks for evaluating foundation models often fall short in providing an accurate assessment of their general abilities in handling human-level tasks. 
This is primarily due to the use of artificial datasets and a lack of emphasis on real-world tasks that require human-like cognitive capabilities.
Moreover, these benchmarks often focus on tasks that do not truly represent the complexities and nuances of real-world human cognition and decision-making, leading to a skewed evaluation of models' capabilities and limiting their ability to provide meaningful insights into the models' real-world applicability.
Consequently, there is a growing need for a more human-centric benchmark that allows for a robust evaluation of foundation model in the context of tasks that are relevant to human reasoning and problem-solving.

In this paper, we introduce a human-centric benchmark \benchmarkname~specifically designed to evaluate the general abilities of foundation models in tasks pertinent to human cognition and problem-solving. 
This benchmark is derived from official, public, and high-standard admission and qualification exams intended for general human test-takers, such as general college admission tests (e.g., Chinese College Entrance Exam (Gaokao) and American SAT), law school admission tests, math competitions, lawyer qualification tests, and national civil service exams. Furthermore, these exams are taken by a diverse range of individuals seeking entry into higher education institutions or new career paths, with millions participating annually (e.g., 12 million for the Chinese Gaokao and 1.7 million for the American SAT). As a result, these exams establish officially recognized standards for assessing human-level capabilities. 
Additionally, the benchmark covers bilingual tasks in both Chinese and English, allowing for a more comprehensive evaluation of the models' capabilities.
By concentrating on these tasks, our benchmark provides a more meaningful and comprehensive evaluation of large language model performance in scenarios directly relevant to human decision-making.
It is worth noting that since state-of-the-art foundation models, such as Text-Davinci-003, ChatGPT, and GPT-4, only have publicly available APIs for language-only tasks, we release the language-only version of \benchmarkname~and focus on evaluating large language models in the present paper. 
Finally, the released \benchmarkname~includes 20 human-centric tasks across a wide variety of subjects.

\begin{figure*}[t]
    \centering
    \includegraphics[width=0.9\textwidth]{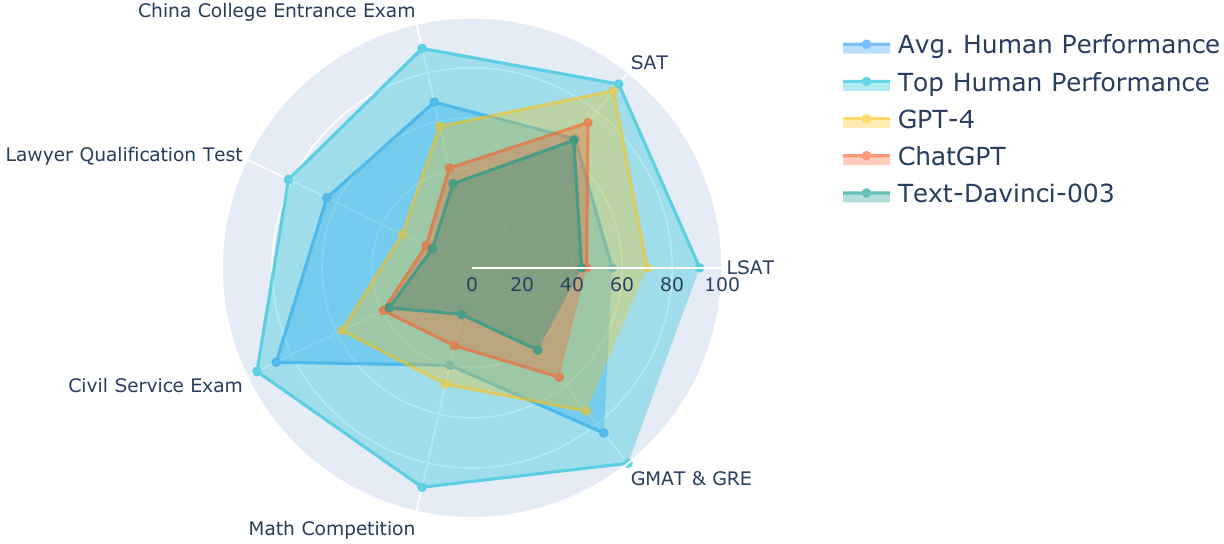}
    \caption{The performance of foundation models (Text-Davinci-003, ChatGPT, and GPT-4) was evaluated on several human-centric exams under zero-shot learning with a Chain-of-Thought (CoT) prompting setting. Human performance (avg.) refers to the average performance of all test takers, while human performance (top) refers to the performance of the top 1\% of test takers, except for the lawyer qualification test which uses the top 10\%. Our results show that, compared to the averaged human performance, GPT-4 achieves better scores on the SAT, LSAT, and math competitions.}
    \label{fig:overview}
\end{figure*}
We employ 20 tasks in our newly-developed benchmark to assess the performance of cutting-edge foundation models, encompassing close-source models, i.e., GPT-4, ChatGPT and Text-Davinci-003, and an open-source model, Vicuna \citep{vicuna2023}.  
Our experiments explore their performance under various settings, including few-shot learning, zero-shot learning, and chain-of-thought prompting techniques. We compare the performance of these models with human performance, as illustrated in Fig. \ref{fig:overview}. 
Remarkably, the results reveal that GPT-4 outperforms the average human performance on LSAT, SAT, and math competitions under the zero-shot chain-of-thought (CoT) setting, demonstrating its capability on human-centric tasks. 
However, there remains a gap between GPT-4 and the top human performance, indicating opportunities for future improvement.
We also discover that these models struggle with tasks requiring complex reasoning (e.g., LSAT-analytical reasoning and physics) or specific domain knowledge, such as law and chemistry.  
Moreover, our comprehensive qualitative analyses of the four dimensions of model capabilities (i.e., \textit{understanding, knowledge, reasoning, and calculation}) delve into their respective strengths and limitations, providing valuable insights into their general capabilities.
This multi-faceted approach enables us to thoroughly examine the models' single-task behavior and identify general patterns, ultimately contributing to a more robust understanding of these state-of-the-art models and their potential applications in tackling human-level tasks.

We aim to drive innovation in developing more effective and reliable AI assistants that advance towards Artificial General Intelligence (AGI). By identifying areas for improvement and understanding their limitations, we can enhance the models' performance and gain a deeper understanding of their underlying mechanisms. Ultimately, this will lead to more powerful and dependable AI systems that better serve the needs of users across a wide range of applications. 

In summary, our research underscores the importance of evaluating foundation models in the context of human-level tasks and provides a solid benchmark for such evaluations. We hope our findings inspire further innovation and progress in the development of large foundation models, ultimately leading to more reliable and effective AI systems.

\section{Background and Related Work}
\subsection{Large Foundation Model}
In recent years, large foundation models, like Large Language Models (LLMs) (e.g., GPT-3~\citep{brown2020language}, GPT-4~\citep{openai2023gpt4}, OPT~\citep{zhang2022opt} and FLAN-T5~\citep{chung2022scaling}) have successfully demonstrated unprecedented performance in a wide range of natural language tasks.
The success of these models can be attributed to advances in deep learning techniques, architectural improvements, and the availability of massive amounts of data for training. 
The most recent cutting-edge language models, such as GPT-4 \citep{openai2023gpt4} and ChatGPT\footnote{\url{https://chat.openai.com/chat}}, have continued to demonstrate substantial adaptability to a diverse array of tasks and domains and have served as a daily decision-making assistant for human beings.

 However, despite their impressive performance on various benchmarks, concerns have been raised about the reasoning abilities, trustfulness and real-world applicability of these models \citep{marcus2019rebooting}.
 Therefore, there is still a need for more comprehensive benchmarks that assess the general reasoning abilities of large foundation models in human-centric tasks. 
\subsection{Existing Benchmarks}
Constructing benchmarks is a reliable way to establish evaluation standards and monitor model performance. Numerous benchmarks~\citep{thorne-etal-2018-fever,rajpurkar2016squad,bowman-etal-2015-large,ehrmann_introducing_2022} have been proposed and widely adopted for evaluating single-task performance, such as SQuAD~\citep{rajpurkar2016squad} for assessing answer extraction ability and SNLI~\citep{bowman-etal-2015-large} for evaluating natural language inference capability.

The emergence of general language models (LMs) like BERT~\citep{devlin-etal-2019-bert} has made it increasingly essential to develop more comprehensive benchmarks to assess the general capabilities of these LMs. GLUE~\citep{wang-etal-2018-glue} and SuperGLUE~\citep{wang2019superglue} are popular benchmarks that evaluate language model performance across diverse NLP tasks, including paraphrase identification, sentiment analysis, and more. GLUE series benchmarks have significantly influenced language model development, encouraging researchers to enhance their models' generalization capabilities. The LAMBADA language modeling task \citep{paperno2016lambada} assesses language models' ability to capture long-range dependencies in text. SentEval~\citep{conneau-kiela-2018-senteval} and DecaNLP~\citep{mccann2018natural} also set benchmarks for evaluating models' general capabilities.

Despite their broad applicability, these benchmarks mainly consist of artificially curated datasets designed to evaluate specific machine skills, rather than real-world problems aimed at assessing human behaviors. Consequently, these benchmarks primarily focus on simpler textual understanding rather than complex reasoning abilities aligned with real-world applicability.

MMLU~\citep{hendrycks2020measuring} addresses this issue by collecting questions from online sources covering a diverse set of subjects (e.g., history, humanities) that humans learn, pushing towards human-centric evaluation. However, there are key differences between MMLU and our work: (1) Source of benchmark: Our benchmark is derived from high-standard and official human-centric exams, such as college admission tests and professional qualification tests, ensuring a robust and well-standardized evaluation of language models. In contrast, MMLU's data source is not explicitly mentioned, and it is unclear whether the tasks come from similarly professional and high-quality sources. (2) Bilingual Evaluation: Our benchmark is bilingual, containing both English and Chinese tasks, which allows for a more comprehensive assessment of language models across different languages and cultures. In contrast, MMLU only contains English data, limiting its evaluation scope to a single language.

\subsection{Evaluation of Recent Large Foundation Models} The rapid development of foundation models has significantly propelled research efforts in evaluating their performance and behavior. As these models continue to grow in size and complexity, it becomes increasingly important to assess their abilities in understanding textual data and performing complex reasoning and problem-solving tasks that humans typically excel at. Consequently, studies have been conducted to understand their strengths and limitations.

ToxiGen~\citep{hartvigsen2022toxigen} and BOLD~\citep{dhamala2021bold} evaluates the bias in language models. 
\citet{liang2022holistic} conducted a holistic evaluation of dozens of language models (up to InstructGPT~\citep{ouyang2022training}) on several previous NLP tasks (e.g., Openbook QA \citep{kwiatkowski-etal-2019-natural,Mihaylov2018CanAS} and text classification \citep{clark2019boolq}), focusing on model behavior analysis. However, this work does not cover the most recent LLMs such as \textit{ChatGPT} and \textit{GPT-4}, and evaluations are mainly conducted on artificially curated NLP datasets rather than real-world scenarios for testing humans.

Recognizing this limitation, recent reports have emphasized the importance of evaluations in human-centric scenarios. For example, \citet{choi2023chatgpt} assessed ChatGPT's performance using essay questions from law schools. 
Recent reports start to investigate the capabilities of GPT-4, like \citet{bubeck2023sparks} and \citet{openai2023gpt4}.
The official technical report of GPT-4~\citep{openai2023gpt4} also underscored the importance of evaluating models' behaviors on human exams and analyzed GPT-4's performance on several such exams. However, the relevant benchmarks in these reports and the corresponding model outputs are not publicly available, and the evaluation metric is also not transparent. 
These factors limit further research to follow up their evaluation and to make direct comparison.

To bridge this gap, we propose \benchmarkname, which releases collected questions from high-standard, official human exams and supports a standardized, automatic evaluation metric. Furthermore, we conduct comprehensive experiments with the most recent LLMs including \textit{ChatGPT} and \textit{GPT-4} and release the model outputs to foster analysis and research within the entire community.

\section{Human-Centric Benchmark}
In this section, we outline the design principles and process of constructing our human-centric benchmark.
\subsection{Design Principles}
\par{\textbf{Emphasis on human-level cognitive tasks:} In designing the human-centric benchmark, our primary objective is to center on tasks that closely align with human cognition and problem-solving, ultimately assessing the general abilities of foundation models in a more meaningful and comprehensive manner. To achieve this, we incorporate a diverse range of official, public, and high-standard admission and qualification exams that cater to general human test-takers. These exams, which include college admission tests, law school admission tests, math exams, lawyer qualification tests, and national civil service exams, are taken by millions of individuals seeking entry into higher education or new career paths. By adhering to these officially recognized standards for evaluating human-level capabilities, our benchmark design ensures that the assessment of model performance is directly relevant to human decision-making and cognitive abilities.}

\par{\textbf{Relevance to real-world scenarios:} The second design principle guiding the construction of our human-centric benchmark is the emphasis on tasks that hold significant implications for real-world situations. By selecting tasks derived from high-standard admission and qualification exams, we ensure that the assessment reflects the complexity and practicality of challenges that individuals commonly encounter in various fields and contexts. This approach allows us to not only gauge the models' performance in relation to human cognitive abilities but also better understand their applicability and effectiveness in real-life scenarios. Consequently, our benchmark design contributes to the development of AI systems that are more reliable, practical, and well-suited for addressing a wide range of real-world problems.}

\subsection{Exam Selection}

In line with the design principles outlined above, we have selected a diverse range of standardized, high-quality exams that emphasize human-level reasoning and real-world relevance. 
Some exams are participated by millions of human test-takers annually. For example, 12 millions of students participate in Gaokao every year. Statistics of annual human participants are reported in Tab. \ref{tab:data-statistic}.
The following categories of human-centric exams are included in our benchmark:
\par{\textbf{General College Entrance Exams:} These exams, include \textbf{Graduate Record Examinations (GRE), Scholastic Assessment Test (SAT) and China College Entrance Exam (Gaokao)}, are designed to assess the general aptitude and subject-specific knowledge of students seeking entry into higher education institutions. College admission tests encompass various subjects and require critical thinking, problem-solving, and analytical skills, making them ideal for evaluating the performance of large language models in relation to human cognition.
More specifically, we collect exams corresponding to 8 subjects from Chinese Gaokao: history, math, English, Chinese, geography, biology, chemistry and physics.  
We select mathematical questions from GRE, select English and math subjects from SAT to construct the benchmark.  
}
\par{\textbf{Law School Admission Test:} Law school admission tests, such as the \textbf{LSAT}, are intended to measure the reasoning and analytical skills of prospective law students. These tests include sections on logical reasoning, reading comprehension, and analytical reasoning, which challenge the test-takers' ability to analyze complex information and draw accurate conclusions. Incorporating these tasks in our benchmark enables us to assess language models' capabilities in legal reasoning and analysis.}
\par{\textbf{Lawyer Qualification Test:} Lawyer qualification tests, such as the bar exam, assess the legal knowledge, analytical skills, and ethical understanding of individuals pursuing a career in law. These exams cover a broad range of legal topics, including constitutional law, contract law, criminal law, and property law, and require candidates to demonstrate their ability to apply legal principles and reason effectively. By incorporating lawyer qualification tests in our benchmark, we can evaluate language models' performance in the context of professional legal expertise and ethical judgment. 
Specifically, we select questions from previous Chinese lawyer qualification tests to build the benchmark.}

\par{\textbf{Graduate Management Admission Test (GMAT)}: The GMAT is a standardized exam designed to assess the analytical, quantitative, verbal, and integrated reasoning skills of prospective graduate business school students. The GMAT consists of sections such as Analytical Writing Assessment, Integrated Reasoning, Quantitative Reasoning, and Verbal Reasoning, which evaluate the test-takers' abilities to think critically, analyze data, and communicate effectively. Incorporating GMAT tasks in our benchmark enables us to assess the language models' performance in a business context and their potential to assist in decision-making and problem-solving in management scenarios.}

\par{\textbf{High School Math Competitions:} High school math competitions, such as the \textbf{American Mathematics Competitions (AMC)} and the \textbf{American Invitational Mathematics Examination (AIME)}, challenge students' mathematical abilities, creativity, and problem-solving skills. These contests cover a wide range of mathematical topics, including number theory, algebra, geometry, and combinatorics, and often feature unconventional problems that require inventive approaches to solve. By incorporating tasks from high school math competitions into our benchmark, we can further evaluate the language models' aptitude for tackling complex and creative mathematical problems, as well as their ability to generate novel solutions.}
\par{\textbf{Chinese Civil Service Examination:} The Chinese Civil Service Examination is a standardized test administered in China to assess the aptitude and skills of individuals seeking entry into the civil service. These exams evaluate a range of competencies, such as general knowledge, reasoning abilities, language skills, and subject-specific expertise related to the roles and responsibilities of various civil service positions in China. Including tasks from national civil service exams in our benchmark allows us to gauge the language models' performance in public administration contexts and their potential to contribute to policy-making, decision-making, and public service delivery processes.}

By including these tasks in our human-centric benchmark, we create a comprehensive assessment that effectively measures the performance of foundation models in various real-world, human-level reasoning scenarios.
\subsection{Construction of Human-Centric Benchmark}
\subsubsection{Data Collection}
As previously mentioned, our human-centric benchmark comprises questions from a diverse range of official and high-quality exams, originally designed for human test-takers. These exams include general college admission tests (GRE, Gaokao, SAT), entrance exams for specific majors (such as LSAT and GMAT), high school math competitions (AMC and AIME), as well as the national civil service examination and lawyer qualification test in China.

Since evaluating model performance on subjective questions is challenging without human expert scoring, we believe such questions are unsuitable for inclusion in this benchmark for consistent assessment. To ensure a robust and standardized evaluation metric, we have removed all subjective questions, retaining only objective ones, such as multiple-choice and fill-in-the-blank questions. 

With regard to data collection, we gather Gaokao and SAT questions from publicly available online sources, along with their corresponding solutions or explanations. For the LSAT, we utilize data from \citet{wang2022lsat} and \citet{zhong-etal-2022-analytical}, which encompasses three tasks (logical reasoning, reading comprehension, and analytical reasoning) from the LSAT administered between 1991 and 2016. For Chinese civil service examinations, we repurpose data from LogiQA \citep{liu2021logiqa}, a dataset built on various types of logical reasoning questions collected from the National Civil Servants Examination of China. It is worth noting that LogiQA consists of bilingual questions (English and Chinese), where the English version is a translated version of the original Chinese version.

For high school math competitions, we employ data from the MATH dataset~ \citep{hendrycks2measuring}, comprising questions from AMC and AIME. Furthermore, we incorporate GRE and GMAT questions from AQaA-RAT~\citep{ling2017program}, which emphasizes algebraic word problems. In the case of the Chinese Civil Service Examination, we reuse instances from JEC-QA~\citep{zhong2019jec}, a large-scale dataset derived from the National Judicial Examination of China. We down-sample the two types of JEC-QA and MATH to 1,000 instances each. 
\begin{table}[htbp]
  \small
  \centering
  \caption{Introduction of the exams included in \benchmarkname. We highlight the number of human participants taking these exams annually (column ``\# Participants"). We also report the number of instances and average token number in \benchmarkname.}
  \resizebox{1.0\linewidth}{!}{
    \begin{tabular}{lllllrr}
    \toprule
    Exams & \#Participants & Language & Tasks & Subject & \multicolumn{1}{l}{\# Instance} & \multicolumn{1}{l}{\#Avg. Token} \\
    \midrule
    \multirow{9}[2]{*}{Gaokao} & \multirow{9}[2]{*}{12M} & \multirow{9}[2]{*}{Chinese} & GK-geography & Geography & 199  & 144 \\
          &       &       & GK-biology & Biology & 210  & 141 \\
          &       &       & GK-history & History & 243  & 116 \\
          &       &       & GK-chemistry & Chemistry & 207  & 113 \\
          &       &       & GK-physics & Physics & 200  & 124 \\
          &       &       & GK-En & English & 306  & 356 \\
          &       &       & GK-Ch & Chinese & 246  & 935 \\
          &       &       & GK-Math-QA & Math  & 351  & 68 \\
          &       &       & GK-Math-Cloze & Math  & 118  & 60 \\
    \midrule
    \multirow{2}[2]{*}{SAT} & \multirow{2}[2]{*}{1.7M } & \multirow{2}[2]{*}{English} & SAT-En. & English & 206  & 656 \\
          &       &       & SAT-Math & Math  & 220  & 54 \\
    \midrule
    \multicolumn{1}{l}{\multirow{2}[2]{*}{\makecell[l]{Lawyer Qualification Test}}} & \multirow{2}[2]{*}{820K} & \multirow{2}[2]{*}{Chinese} & JEC-QA-KD & Law   & 1000  & 146 \\
          &       &       & JEC-QA-CA & Law   & 1000  & 213 \\
    \midrule
    \multicolumn{1}{l}{\multirow{3}[2]{*}{\makecell[l]{Law School \\ Admission Test (LSAT)}}} & \multirow{3}[2]{*}{170K} & \multirow{3}[2]{*}{English} & LSAT-AR & Law-Analytics & 230  & 154 \\
          &       &       & LSAT-LR & Law-Logic & 510  & 178 \\
          &       &       & LSAT-RC & Law-Reading & 260 & 581 \\
    \midrule
    \multicolumn{1}{l}{\multirow{2}[2]{*}{\makecell[l]{Civil Service Examination}}} & 2M    & English & LogiQA-en & Logic & 651  & 144 \\
          & 2M    & Chinese & LogiQA-ch & Logic & 651  & 242 \\
    \midrule
    GRE   & 340K    & English & \multirow{2}[2]{*}{AQuA-RAT} & \multirow{2}[2]{*}{Math} & \multirow{2}[2]{*}{254} & \multirow{2}[2]{*}{77} \\
    GMAT  & 150K    & English &       &       &       &  \\
    \midrule
    AMC   & 300K    & English & \multirow{2}[2]{*}{MATH} & \multirow{2}[2]{*}{Math} & \multirow{2}[2]{*}{1000} & \multirow{2}[2]{*}{40} \\
    AIME  & 3000    & English &       &       &       &  \\
    \bottomrule
    \end{tabular}}%
  \label{tab:data-statistic}%
\end{table}%

As a result, we construct a benchmark consisting of 8,062 questions for evaluation. Detailed data statistics are presented in Table \ref{tab:data-statistic}. It is worth noting that our benchmark is bilingual, encompassing both \textbf{English and Chinese tests}. 
This design enables the evaluation of a broader scope of model capabilities, reflecting their performance and adaptability across different languages. 

\subsubsection{Evaluation Metrics}
The benchmark questions consist of objective formats: multiple-choice and fill-in-the-blank questions. For multiple-choice questions, we adopt standard classification accuracy as the evaluation metric. For fill-in-the-blank questions, we employ Exact Match (EM) and F1 metrics. 

\subsubsection{Human Performance}
It is worth noting that we also report the average and top performances of human test-takers as  \textbf{human-level boundaries} for each task.
The human performance results for MATH, LogiQA, and JEC-QA were gathered from their respective publications, while for AQuA-RAT (GMAT, GRE), LSAT, SAT, and Gaokao, we estimated the human performances by scaling the scores of the average (50\%) and top (1\%) test takers against the full scores to 100. It is important to note that while our datasets and human accuracies provide a useful approximation of human test-taker ability, they do not fully represent the range of skills and knowledge that such individuals may possess.

\section{Evaluation of Foundation Models}
\subsection{Foundation Models Selected for Evaluation}
In this section, we evaluate the performance of various state-of-the-art language models on our benchmark dataset. The selected models represent a diverse range of capabilities and are widely recognized in the field. The language models chosen for evaluation include:

\begin{itemize}
\item \textbf{\textit{GPT-4}}: The fourth-generation model in the GPT series, GPT-4 is a state-of-the-art, large-scale generative pre-trained transformer model with enhanced performance and a broader knowledge base compared to its predecessors. While it does not surpass human capabilities in many real-world scenarios, it demonstrates human-level performance on many scenarios. GPT-4 has been iteratively aligned using insights from adversarial testing and ChatGPT, resulting in remarkable improvements in factuality, 
\item \textbf{\textit{ChatGPT}}: A conversational AI model developed by OpenAI, ChatGPT is designed to engage in interactive and dynamic conversations. It has been trained on a vast instruction dataset and further tuned by reinforcement learning with human feedbacks (RLHF), enabling it to provide contextually relevant and coherent responses aligned with human expectation.
\item \textbf{\textit{Text-Davinci-003}}: GPT-3.5 is an intermediate version between GPT-3 and GPT-4, offering improved performance over GPT-3 through further instruction tuning. 
It serves as a bridge between the two models, allowing for comparative analysis. Specifically, we select the variant \textbf{Text-Davinci-003} from GPT-3.5 series for evaluation.
\item \textbf{Vicuna-13B~\citep{vicuna2023}: }It is an open-source LLM, trained on user-shared conversations from ShareGPT by fine-tuning LLaMA. It achieves over 90\% of the quality of OpenAI's ChatGPT. As of May 2023, Vicuna-13B ranks among the top models on the Open LLM Leaderboard~\citep{open-llm-leaderboard}.

A recent technical report~\citep{InternLM} further evaluates \textbf{foundation models (GLM-130B~\citep{du2022glm}, LLaMa-65B~\citep{touvron2023llama} and InternLM-104B~\citep{InternLM})} on AGIEval.
\end{itemize}

In the following sections, we will discuss the evaluation methodology, including the metrics used to assess the performance of these models, and present a comparative analysis of their performance on the benchmark dataset.

\subsection{Experimental Setup}
\begin{figure*}[h]
    \centering
    \includegraphics[width=\textwidth]{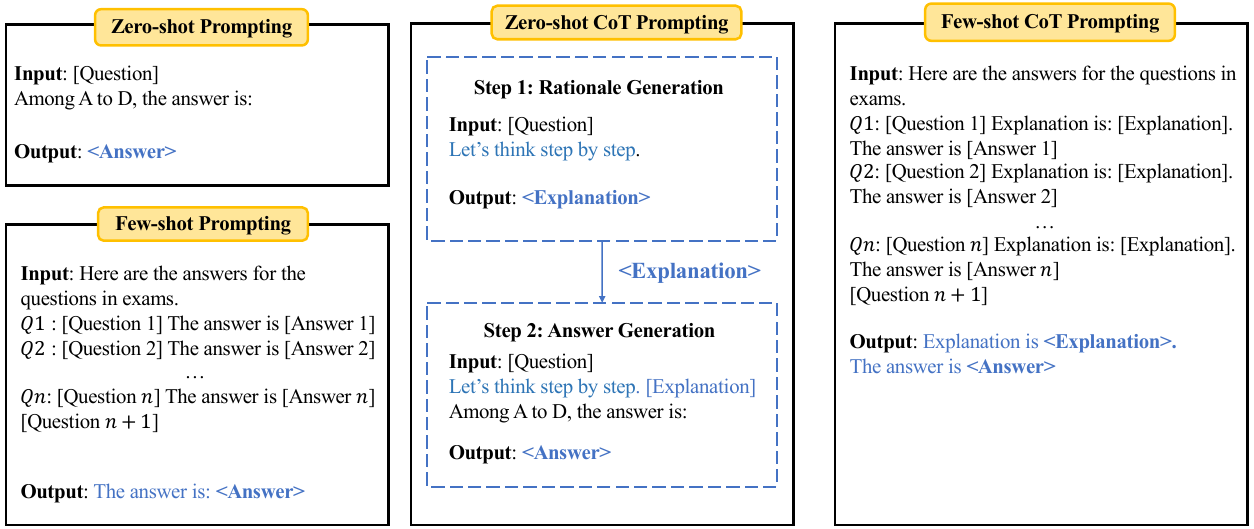}
    \caption{Prompting examples of different settings.}
    \label{fig:prompt-example}
\end{figure*}

In this section, we describe the experimental setup used to evaluate the performance of large language models, including GPT-4, ChatGPT and Text-Davinci-003, on our human-centric reasoning tasks benchmark.
To gauge the adaptability of the language models, we conducted two types of evaluations: zero-shot and few-shot. We further implement a ``Chain-of-Thought'' reasoning evaluation for both zero-shot and few-shot learning. 
Fig. \ref{fig:prompt-example} describes the concrete prompting examples for zero-shot testing, few-shot testing and chain-of-thought prompting.
\textbf{To foster analyses about models' behavior and foster relevant research, we release the model outputs under all the settings.}
\subsubsection{Zero-shot and Few-shot Evaluation}

In the zero-shot setting, models were directly evaluated on the questions without being provided any prior examples of the specific tasks. This scenario tests the models' innate ability to reason and solve problems without explicit training.

In the few-shot setting, models were given a small number of examples (e.g., 5) from the same task before being evaluated on the test samples. This evaluation setup tests the models' ability to quickly adapt and learn from limited examples, simulating real-world scenarios where direct supervision may be scarce.
\subsubsection{Chain-of-Thought (CoT) Reasoning}
Chain-of-Thought (CoT) prompting \citep{wei2022chain} enables large language models to break down a complex question to a series of decomposed reasoning steps.
To further probe the models' reasoning capabilities, we implemented a ``Chain-of-Thought'' reasoning evaluation. This method involves asking the model to first generate an explanation for the given question, and subsequently, answer the question based on its self-generated explanation.

More specifically, as shown in Fig. \ref{fig:prompt-example}, the Chain-of-Thought reasoning evaluation is carried out in two steps for each task in the benchmark.
In the first step, the model is prompted with ``\textit{[question] Let's think step by step: }'' to generate a detailed explanation of its understanding of the problem and the reasoning process required to solve it \citep{zhang2022automatic}.
This allows us to evaluate the model's ability to comprehend complex tasks and identify the essential elements needed for successful problem-solving.
In the second step, the model is prompted with the generated explanation ``\textit{[question] Explanation is: [explanation]. The answer is: }'', and provide an answer to the question, taking into account its self-generated explanation from the first step. 
This process aimed to test the model's ability to use its own reasoning to derive a coherent and accurate solution, simulating the way humans often rely on their understanding and internal reasoning processes to solve problems.
Specially, as shown in Fig. \ref{fig:prompt-example}, in the few-shot CoT setting, the explanation and answer are generated in the same step. 
\subsubsection{Evaluation Metrics}
To assess the performance of the language models on our benchmark, we used both quantitative and qualitative evaluation metrics. Quantitative metrics included task-specific accuracy for multi-choice questions and use Exact Match (EM) for fill-in-blank questions. For a more in-depth analysis, we also performed qualitative evaluations, which involved human evaluators assessing the models' responses in terms of semantic understanding capability, knowledge utilization, and reasoning quality. 
This combination of evaluation metrics allowed us to obtain a comprehensive understanding of the models' strengths and limitations in human-centric reasoning tasks.

\subsubsection{Implementation Details}
All experiments were conducted using the respective language models' API provided by Azure OpenAI Service\footnote{\url{https://azure.microsoft.com/en-us/products/cognitive-services/openai-service}}. The Azure OpenAI services offer two types of APIs: completion and chat completion. The completion API generates text based on prompts, while the chat completion API generates the next AI response based on the conversation history and new human input. For Text-Davinci-003, we use the completion API, and for ChatGPT and GPT-4, we use the chat completion API. Notably, only the chat completion API is available for GPT-4 at present. We use a temperature of zero to generate output using greedy search and set the maximum number of tokens for generation to 2048. Additionally, we set the frequency penalty to zero and top p to 1, which are the default values for these APIs.

The Chat Completion API exhibits distinct properties in comparison to the Completion API. In a zero-shot context, the Chat Completion API has the potential to autonomously generate reasoning steps, eliminating the necessity for prompt engineering and potentially enhancing performance. For few-shot scenarios, it is imperative to adapt the few-shot examples into conversational history, as recommended in the Azure guidelines. The inquiry is transformed into a user input, while the AI's response is composed of a chain-of-thought explanation and answer. However, we have observed that the models, particularly ChatGPT, encounter difficulties in adhering to the pattern using the Chat Completion API. Consequently, we employ the Completion API to conduct few-shot experiments with ChatGPT, which is analogous to Text-Davinci-003, in order to gain a deeper understanding of the disparities between Text-Davinci-003 and ChatGPT. If a completion API for GPT-4 become accessible in the future, we will revise and update the few-shot outcomes accordingly.

\paragraph{Few-shot  Examples Construction:}
For AQuA-RAT, LogiQA and LSAT, we randomly sample five examples of medium sentence length of the test set from the provided training set. Similarly, for Gaokao and SAT, we randomly select five examples of medium sentence length from the dataset that was initially collected and exclude them from the test set. For JEC-QA, given that the test set is not publicly available, we take the first 1,000 examples from the training set as the test set and again sample five examples of medium sentence length from the rest. For MATH, we use the same instances as in the appendices of \citet{lewkowycz2022solving}.

To generate explanations for few-shot CoT experiments, for AQuA-RAT and MATH, we use the existing rationales from these datasets.
For Gaokao and SAT, we collected expert annotations. For LogiQA, JEC-QA and LSAT, we use ChatGPT to generate explanations given the questions and the answers. We release all CoT demonstrations in the Github repository.
\begin{table}[htbp]
  \small
  \centering
  \caption{Performance of LLMs on 20 tasks under \textbf{zero-shot} and \textbf{zero-shot CoT} settings. We also report human performance on each task. For LSAT, Gaokao and SAT, we report average (50\%) and top (1\%) human performance. The Text-Davinci-003 is abbreviated as TD-003.}
  
  \resizebox{1.0\linewidth}{!}{
    \begin{tabular}{l|cc|ccc|ccc}
    \toprule
          & \multicolumn{2}{c|}{Human Performance} & \multicolumn{3}{c|}{Zero-Shot } &       & Zero-Shot CoT &  \\
    \midrule
    Task/Model & Avg.  & Top   & TD-003 & ChatGPT & GPT-4 & TD-003 & ChatGPT & GPT-4 \\
    \midrule
    AQuA-RAT & 85    & 100   & 29.9  & 31.9  & 40.6  & 42.1  & 55.9  & 73.2 \\
    MATH  & 40    & 90    & 11.9  & 26.4  & 35.7  & 19.1  & 31.9  & 47.7 \\
    LogiQA (English) & 86    & 95    & 22.7  & 35.0    & 49.3  & 36.9  & 39.9  & 57.8 \\
    LogiQA (Chinese) & 88    & 96    & 40.3  & 41.0    & 58.8  & 36.7  & 38.9  & 57.5 \\
    JEC-QA-KD & 71    & 78    & 21.9  & 21.1  & 33.4  & 18.4  & 21.2  & 31.9 \\
    JEC-QA-CA & 58    & 85    & 21.0  & 22.0    & 31.1  & 16.7  & 19.6  & 29.8 \\
    LSAT-AR & 56    & 91    & 21.7  & 24.4  & 35.2  & 23.9  & 22.6  & 34.4 \\
    LSAT-LR & 56    & 91    & 47.5  & 52.6  & 80.6  & 50.0    & 52.6  & 80.6 \\
    LSAT-RC & 56    & 91    & 64.7  & 65.4  & 85.9  & 57.6  & 62.1  & 85.1 \\
    SAT-Math & 66    & 94    & 35.5  & 42.7  & 64.6  & 54.6  & 70.9  & 95.0 \\
    SAT-English & 66    & 94    & 74.8  & 81.1  & 88.8  & 75.7  & 77.7  & 85.9 \\
    SAT-English (w/o Psg.) & 66    & 94    & 38.4  & 44.2  & 51.0    & 44.2  & 45.6  & 25.2 \\
    GK-Cn & 65    & 85    & 43.9  & 39.0    & 53.3  & 35.4  & 33.7  & 44.7 \\
    GK-En & 69    & 91    & 81.4  & 84.9  & 91.9  & 83.0    & 84.3  & 92.5 \\
    GK-geography & 65    & 85    & 53.3  & 59.8  & 76.9  & 48.7  & 55.8  & 72.4 \\
    GK-history & 64    & 85    & 47.3  & 59.7  & 77.4  & 37.0    & 50.2  & 76.5 \\
    GK-biology & 68    & 89    & 40.5  & 52.9  & 75.7  & 30.0    & 42.4  & 71.9 \\
    GK-chemistry & 66    & 86    & 27.1  & 38.7  & 51.7  & 24.6  & 33.8  & 52.2 \\
    GK-physics & 71    & 94    & 22.0  & 33.0    & 39.0    & 18.5  & 29.5  & 45.5 \\
    GK-Math-QA & 73    & 96    & 28.2  & 36.5  & 47.0    & 28.8  & 33.3  & 50.7 \\
    GK-Math-Cloze & 73    & 96    & 17.0  & 7.6   & 16.1  & 4.2   & 5.1   & 15.3 \\
    \midrule
    Average & 67    & 91    & 38.1  & 42.9  & 56.4  & 37.4 & 43.2  & 58.4 \\
    \bottomrule
    \end{tabular}
    }
    %
   
  \label{tab:zero-shot}%
\end{table}%

\begin{table}[htbp]
  \small
  \centering
  \caption{Performance of LLMs on 20 tasks under \textbf{few-shot} and \textbf{few-shot CoT} settings. We also report human performance on each task. For LSAT, Gaokao and SAT, we report average (50\%) and top (1\%) human performance. The Text-Davinci-003 is abbreviated as TD-003.}
  \resizebox{1.0\linewidth}{!}{
    \begin{tabular}{l|cc|ccc|ccc}
    \toprule
          & \multicolumn{2}{c|}{Human Performance} & \multicolumn{3}{c|}{Few-Shot } &       & Few-Shot CoT &  \\
    \midrule
    Task/Model & Avg.  & Top   & TD-003 & ChatGPT & GPT-4 & TD-003 & ChatGPT & GPT-4 \\
    \midrule
    AQuA-RAT & 85    & 100   & 30.3  & 31.1  & 50.8  & 47.2  & 60.6  & 74.0 \\
    MATH  & 40    & 90    & 10.3  & 14.8  & 15.1  & 15.1  & 30.1  & 25.3 \\
    LogiQA (English) & 86    & 95    & 43.5  & 43.5  & 63.9  & 37.5  & 38.9  & 62.7 \\
    LogiQA (Chinese) & 88    & 96    & 43.2  & 46.2  & 65.0    & 40.0    & 38.6  & 61.9 \\
    JEC-QA-KD & 71    & 78    & 22.4  & 27.6  & 41.3  & 23.6  & 23.4  & 40.4 \\
    JEC-QA-CA & 58    & 85    & 22.2  & 25.1  & 37.4  & 16.1  & 20.0    & 34.7 \\
    LSAT-AR & 56    & 91    & 22.6  & 25.7  & 33.9  & 22.6  & 25.2  & 31.7 \\
    LSAT-LR & 56    & 91    & 60.4  & 59.2  & 85.9  & 51.2  & 52.2  & 84.5 \\
    LSAT-RC & 56    & 91    & 70.6  & 67.7  & 87.7  & 64.3  & 57.6  & 87.7 \\
    SAT-Math & 66    & 94    & 44.6  & 40.9  & 71.8  & 55.5  & 65.0    & 89.6 \\
    SAT-English & 66    & 94    & 84.0    & 81.1  & 88.8  & 76.7  & 78.2  & 85.9 \\
    SAT-English (w/o Psg.) & 66    & 94    & 48.1  & 53.9  & 63.6  & 48.5  & 51.5  & 62.6 \\
    GK-Cn & 65    & 85    & 25.6  & 41.5  & 61.4  & 29.3  & 37.8  & 51.6 \\
    GK-En & 69    & 91    & 86.9  & 86.3  & 93.8  & 80.7  & 84.6  & 93.1 \\
    GK-geography & 65    & 85    & 59.8  & 63.8  & 75.9  & 52.3  & 61.8  & 76.4 \\
    GK-history & 64    & 85    & 49.0    & 57.6  & 77.8  & 51.9  & 58.4  & 78.2 \\
    GK-biology & 68    & 89    & 44.3  & 52.4  & 80.0    & 32.9  & 50.0    & 72.9 \\
    GK-chemistry & 66    & 86    & 32.4  & 44.0    & 54.6  & 35.8  & 33.8  & 54.1 \\
    GK-physics & 71    & 94    & 31.0    & 33.5  & 43.5  & 27.5  & 36.5  & 54.5 \\
    GK-Math-QA & 73    & 96    & 27.6  & 31.3  & 39.9  & 33.1  & 31.6  & 49.0 \\
    GK-Math-Cloze & 73    & 96    & 5.9   & 5.9   & 11.0    & 5.93  & 8.5   & 16.1 \\
    \midrule
    Average & 67    & 91    & 41.2 & 44.4  & 59.2  & 40.4 & 45    & 61.3 \\
    \bottomrule
    \end{tabular}}%
  \label{tab:few-shot}%
\end{table}%

\begin{table}[htbp]
  \centering
   \setlength{\tabcolsep}{4pt}
  \caption{Performance of open-source model Vicula-13B under zero-shot and zero-shot CoT setting. Task names are abbreviated.}
  \resizebox{1.0\linewidth}{!}{
    \begin{tabular}{l|cc|cc|cc|ccc|cc|ccccccccc}
    \toprule
    \multicolumn{1}{c|}{\multirow{2}[4]{*}{Task/Model}} &       &       & \multicolumn{2}{c|}{LogiQA} & \multicolumn{2}{c|}{JEC-QA} &       & \multicolumn{1}{l}{LSAT} &       & \multicolumn{2}{c|}{SAT} & \multicolumn{9}{c}{GK} \\
\cmidrule{2-21}          & \multicolumn{1}{l}{AQaA} & \multicolumn{1}{l|}{MATH} & \multicolumn{1}{l}{En.} & \multicolumn{1}{l|}{Cn.} & \multicolumn{1}{l}{KD} & \multicolumn{1}{l|}{CA} & \multicolumn{1}{l}{AR} & \multicolumn{1}{l}{LR} & \multicolumn{1}{l|}{RC} & \multicolumn{1}{l}{Math} & \multicolumn{1}{l|}{En.} & \multicolumn{1}{l}{Cn} & \multicolumn{1}{l}{En} & \multicolumn{1}{l}{Geo.} & \multicolumn{1}{l}{His.} & \multicolumn{1}{l}{Bio.} & \multicolumn{1}{l}{Che.} & \multicolumn{1}{l}{Phy.} & \multicolumn{1}{l}{M.-QA} & \multicolumn{1}{l}{M.-Cloze} \\
    \midrule
    Vicuna (ZS) & 26.4  & 6.8   & 18.4  & 23.5  & 14.3  & 12.4  & 22.2  & 25.5  & 30.5  & 24.6  & 50.5  & 25.6  & 50.7  & 24.6  & 28.9  & 20.5  & 26.6  & 15    & 22.5  & 2.5 \\
    Vicuna (ZS-CoT) & 22.1  & 6.6   & 30.3  & 27.1  & 14.9  & 15.2  & 20.9  & 36.1  & 44.2  & 35.5  & 57.8  & 23.6  & 67    & 28.6  & 34.9  & 24.3  & 23.2  & 17    & 21.7  & 1.7 \\
    \bottomrule
    \end{tabular}}%
  \label{tab:open-source-result}%
  \vspace{-0.15in}
\end{table}%
\subsection{Main Results}
The results of \textbf{zero-shot learning and zero-shot CoT (Chain-of-Thought)} are reported in Tab. \ref{tab:zero-shot}.
The results of \textbf{few-shot learning and few-shot CoT} are reported in Tab. \ref{tab:few-shot}.
We also report average and top human performance on each task. From the results, we highlight the following findings.
\par{\textbf{(1) Superior Performance of GPT-4:}}
Our results indicate that, on average, GPT-4 significantly outperforms its counterparts (ChatGPT and Text-Davinci-003) across all four evaluation settings. 
Impressively, GPT-4 achieves 93.8\% accuracy on Gaokao-English and 95\%  accuracy on SAT-MATH, demonstrating its superior general capabilities 
 in handling human-centric tasks.
  \par{\textbf{(2) Comparison between ChatGPT and Text-Davinci-003:}}
 Our analysis shows that ChatGPT significantly outperforms  Text-Davinci-003 in tasks that require a high degree of external knowledge, such as those involving geography, biology, chemistry, physics, and mathematics. This suggests that ChatGPT has a stronger knowledge base and is better equipped to handle tasks that necessitate a deep understanding of specific domains.

On the other hand, ChatGPT slightly outperforms Text-Davinci-003 or achieves comparable results in tasks that require pure understanding and do not rely heavily on external knowledge, such as English and LSAT tasks, across all evaluation settings. This observation implies that both models are capable of handling tasks centered on language comprehension and logical reasoning without the need for specialized domain knowledge.
 \par{\textbf{(3) Challenge of Complex Tasks:}}
 Despite the overall good performance of the models, we observe that all the LLMs struggle with complex reasoning tasks, such as MATH, LSAT-AR, GK-physics, and GK-Math, across all evaluation settings. This highlights the limitations of these models in handling tasks that require advanced reasoning and problem-solving skills. The observed difficulties in tackling complex reasoning problems present an opportunity for future research and development, aiming to enhance the models' general reasoning capabilities.
 \par{\textbf{(4) Few-shot Learning vs. Zero-shot Learning:}}
 In our experiments, we observe that few-shot learning generally leads to only a limited improvement in performance compared to zero-shot learning. This finding suggests that the zero-shot capabilities of current large language models (LLMs) are approaching their few-shot learning abilities. This marks a significant advancement compared to the original GPT-3 \citep{brown2020language} model, where few-shot performance was considerably better than zero-shot.
One plausible explanation for this development is the enhanced human-alignment and instruction tuning in current LLMs. These improvements enable the models to better understand the meaning and context of tasks in advance, thus allowing them to perform well even in zero-shot settings. This progress in LLMs' zero-shot capabilities highlights the effectiveness of recent advancements in instruction tuning of LLMs.

As shown in Fig. \ref{tab:open-source-result}, Vicuna, despite excelling on OpenLLM leaderboard~\citep{open-llm-leaderboard} and its claimed comparable ability with ChatGPT, falls short on AGIEval, highlighting the valuable challenges our benchmark presents to open-source models.
\subsection{Analyses of Chain-of-thought Prompting (CoT)}
\label{sec:cot-analysis}
As reported in Tab. \ref{tab:zero-shot} and Tab. \ref{tab:few-shot}, the Chain-of-Thought (CoT) prompting technique demonstrates its potential by improving average zero-shot and few-shot performance. However, the performance gains from CoT are not consistently observed across all tasks. Our analysis of CoT leads to the following findings:
\par
\textbf{(1) Performance Variability:} CoT substantially enhances performance in English mathematical exams, including MATH, AQuA-RAT, and SAT-Math. However, it leads to performance degradation in several other tasks, which may be a consequence of the model generating misleading reasoning processes. We also observed that the effects of CoT vary across different tasks, indicating that its impact on model performance is not uniform. It is crucial to further investigate the factors contributing to these performance variations and identify ways to optimize CoT for a broader range of tasks. 
\par
\textbf{(2) Backbone Dependency:} The effectiveness of CoT is influenced by the underlying model. For example, GPT-4 is better equipped to generate illustrative reasoning processes, which subsequently result in enhanced performance when employing CoT. This observation highlights the importance of considering the interplay between CoT and the backbone model, as their compatibility can significantly impact problem-solving capabilities.
\par
\textbf{(3) Language Sensitivity:} The impact of CoT varies across different languages. In the case of the LogiQA exam, the English version is translated from the original Chinese version. Both ChatGPT and GPT-4 demonstrate performance improvements with CoT in LogiQA-English but suffer from performance degradation in LogiQA-Chinese. A similar pattern is observed in mathematical tests, where CoT boosts performance in English math exams (MATH, AQuA) but leads to performance declines in the Chinese math exam in Gaokao.
These findings suggest that CoT's effectiveness is sensitive to language variations, emphasizing the need to further generalize and optimize CoT for different linguistic contexts. By tailoring CoT to better accommodate diverse languages, we can ensure more consistent and reliable problem-solving capabilities across a wider range of tasks and language settings.

These observations indicate that the effectiveness of CoT is relevant to task, model capability, and involved language. These factors should be carefully considered when employing CoT for specific tasks or developing future language models.

\subsection{Qualitative Analyses of Model Capabilities}

To gain a deeper understanding of the models' alignment with human capabilities, we perform a qualitative analysis by sampling 100 errorly answered instances for each task. Specifically, we examine the outputs from ChatGPT under the zero-shot CoT setting, where the model is required to generate both explanations and the corresponding answers. This approach allows us to assess the model's ability to handle the reasoning process and answer questions without further supervision.

\begin{figure*}[t]
    \centering
    \includegraphics[width=\textwidth]{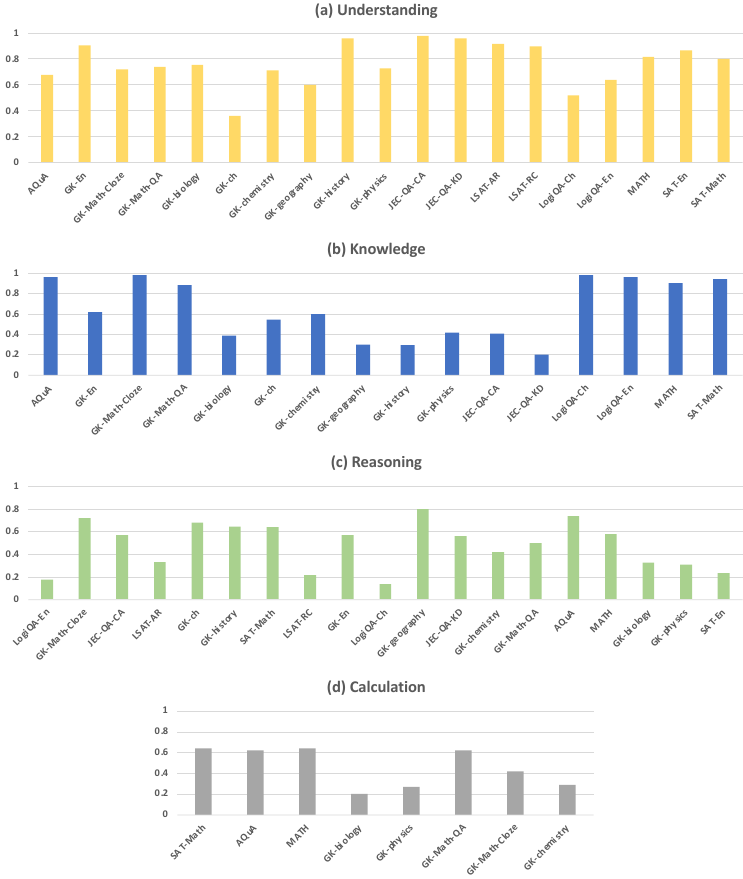}
    \caption{Qualitative assessment of inaccurately answered questions by the model focuses on four dimensions of capabilities: understanding, knowledge acquisition, reasoning and calculation.}
    \label{fig:ability-analysis}
    \vspace{-2mm}
\end{figure*}

We enlist human annotators with expert knowledge, such as Ph.D. students and professional researchers, to evaluate the model outputs (i.e., explanations and answers) along the following four dimensions. The average scores on tasks for the four dimensions of capabilities are shown in Fig. \ref{fig:ability-analysis}. 
\par
\textbf{Understanding:} Evaluating whether the model can accurately comprehend the semantic meaning of the context and questions.
\par
\textbf{Knowledge:} Assessing whether the model can accurately recall relevant external knowledge or formulas for problem-solving.
\par
\textbf{Reasoning:} Determining whether the model can perform correct reasoning steps or formulate accurate reasoning ideas to solve problems.
\par
\textbf{Calculation:} Assessing whether the model can make correct mathematical calculations of the given formulas in the contexts of math, biology, chemistry and physics.

An instance is scored with 1 if the model performs correctly on the corresponding skill and 0 otherwise.
It is worth noting that some tasks, such as LSAT and certain English tasks, primarily emphasize understanding and reasoning without requiring extensive external knowledge or calculation. These tasks are excluded from the corresponding skill analyses to maintain focus on the relevant capabilities.

By conducting this qualitative analysis, we can obtain a more detailed understanding of the models' strengths and weaknesses in various aspects, shedding light on areas that may require further improvement in future iterations of large language models.

In addition to the capability scoring, we request annotators to provide a summary of the common detailed insights regarding the models' behavior patterns, emphasizing their strengths and weaknesses in addressing these human-centric tasks. We analyze the overall trends of the four aforementioned dimensions of capabilities in Section \ref{sec:model-ability-trend}, outline the models' strengths in Section \ref{sec:strength}, and discuss their weaknesses in Section \ref{sec:weakness}.
\subsubsection{Overall Trend of Model Capabilities}
\label{sec:model-ability-trend}
From the qualitative analysis in Fig. \ref{fig:ability-analysis}, we have the following observations:

\textbf{Understanding:} The model generally performs well in the understanding dimension. For most tasks, it can accurately interpret the meaning of questions, demonstrating its ability to comprehend context effectively. 

\textbf{Knowledge:} In the knowledge dimension, the model demonstrates proficiency in identifying correct knowledge or formulas for mathematical and logical tasks. However, it encounters difficulties in recalling specific domain knowledge, such as law, biology, and physics. This observation emphasizes the significance of integrating more domain-specific knowledge into the model, potentially through the utilization of specialized domain-specific knowledge bases or knowledge-enhanced pre-training techniques.

\textbf{Reasoning:} Among the four dimensions, the model's reasoning capability appears to be relatively underdeveloped. For tasks necessitating complex, multi-step reasoning (e.g., LSAT-AR, LogiQA, and GK-Physics), the model struggles to accurately execute multi-step reasoning processes. This underlines the importance of future research concentrating on augmenting the model's reasoning capabilities, potentially through the exploration of innovative prompting methods or training strategies that encourage complex reasoning and problem-solving skills.

\textbf{Calculation}: The model's calculation ability is weaker than their understanding capacity and displays variability across different subjects. They perform better in math exams, but face challenges in chemistry and biology exams, which often require frequent variable substitution involving chemical elements. This suggests that enhancing the calculation and combinatorial abstraction and calculation ability of the model, particularly in subject areas with specialized notations or customized symbol substitutions, is a crucial challenge for further improvement.

These insights into the model's performance across the four dimensions provide valuable information to guide future research and development efforts aimed at enhancing large language models' capabilities in addressing human-centric tasks.

\subsubsection{Strengths}
\label{sec:strength}
By closely examining the models' output explanations and analyzing their behavior patterns, we identify several strengths that highlight the capabilities of these models in handling various aspects of problem-solving. The models demonstrate remarkable performance in the following areas:
\par
\textbf{Good Understanding:} The models excel in accurately comprehending the semantic meaning of context and questions. They effectively discern nuances, interpret complex questions, and parse intricate sentences, showcasing their strong natural language understanding skills. This capability enables them to grasp the core concepts of a problem and lays the foundation for subsequent reasoning and problem-solving steps.
\par
\textbf{Proficiency in Simple Reasoning and Deduction:} The models are adept at handling tasks that require simple reasoning and deduction. They can draw straightforward conclusions, identify logical connections, and perform basic inference, which is crucial for addressing a wide variety of problems. Their ability to effectively perform simple reasoning tasks is an essential component of their overall problem-solving skillset.

For example, the model can comprehend ``could be true except'' is equals to ``cannot be true''.
Also, taking a question in the LSAT-AR task as an example, it requires the model to place 8 books to a bookcase with three shelves following conditions: \textit{``each shelf should have at least 2 books and more books should be placed on the bottom shelf than on the top shelf''}. The model successfully deduced that ``\textit{there are at least 3 books on the bottom shelf and at most 2 books on the top shelf.}''
\par 
\textbf{Grasping General Reasoning Process:} The models demonstrate an ability to understand and generate the general idea of reasoning processes. They can identify the main components of a problem, recognize the structure of a solution, and outline a high-level reasoning strategy. This capability allows them to generate meaningful explanations and provides a starting point for more detailed reasoning and problem-solving tasks.

These strengths indicate that the models have made significant progress in aligning with human problem-solving capabilities. However, there is still room for improvement, especially in complex reasoning tasks and domain-specific knowledge, as discussed in the subsequent section on weaknesses.

\subsubsection{Weaknesses}
\label{sec:weakness}

Despite the significant strengths displayed by the models, there are certain limitations that need to be addressed to improve their overall performance. We outline these weaknesses based on the analysis of the models' output explanations:

\textbf{Understanding:}
\begin{itemize}
\item \textit{Difficulty with Variable Substitution:} The models struggle to understand questions that require variable substitution, often failing to recognize the need for this operation and how it should be applied to solve the problem. This limitation can hinder their ability to tackle a wide range of mathematical and logical tasks. 
For instance, the model frequently struggles to answer chemistry questions that involve substituting a variable in a chemical equation with a chemical element and analyzing its properties.
\item \textit{Challenges with Complex Math Concepts and Symbols:} The models find it difficult to comprehend complex mathematical concepts and interpret the meaning of symbols, particularly when multiple symbols are involved. This weakness limits their ability to effectively address advanced mathematical problems.
\item \textit{Confusion with Similar Concepts:} The models can easily be confused by similar concepts or terms, sometimes leading to incorrect or misleading reasoning. 
For example, in the physics exam, the model is confused by the difference between vertical speed and horizontal speed of moving object.
This issue underscores the need for better disambiguation and concept understanding techniques in future model iterations.

\item \textit{Difficulty in Handling Long Contexts:} The models are prone to being disrupted by long contexts, leading to a decline in their comprehension and reasoning abilities. Improving the models' capacity to maintain focus and process extensive information is essential for enhancing their performance in real-world scenarios.

\end{itemize}
\textbf{Knowledge:}
\begin{itemize}
\item \textit{Insufficiency in Commonsense and Domain-Specific Knowledge:} The models occasionally demonstrate a lack of commonsense or domain-specific knowledge, which hinders their ability to generate plausible explanations and provide accurate answers. This limitation underscores the importance of incorporating diverse knowledge sources into the training data and exploring techniques that can more effectively integrate and access this information within the models. Moreover, it emphasizes the necessity to broaden the models' exposure to a wider array of subjects and fields, ensuring a more comprehensive understanding of various domains.

For instance, given the conditions ``\textit{if Julio and Kevin both lead morning sessions, we know that Kevin and Rebecca must lead sessions that meet on the same day},'' the model incorrectly deduces that ``\textit{Therefore, Rebecca must also lead a morning session.}'' This indicates a lack of commonsense knowledge about the relationship between \textit{morning} and \textit{day}, leading to an erroneous explanation. Additionally, the model generally performs poorly on tasks requiring specific domain knowledge, such as law and chemistry.

\item \textit{Difficulty Identifying Correct Formulas:} The models occasionally struggle to recall and apply the appropriate formulas necessary to solve particular problems, especially in tasks that demand specialized knowledge or expertise. This shortcoming suggests that there is potential for improvement in the models' knowledge retrieval mechanisms and their ability to recognize the relevance of specific formulas to a given problem. Developing strategies to enhance the models' proficiency in identifying and applying correct formulas will be essential for improving their performance in tasks requiring a deep understanding of domain-specific concepts and techniques.
\end{itemize}

Addressing these weaknesses in knowledge will contribute to the development of more robust and versatile large language models, better equipped to tackle a broader range of human-centric tasks and exhibit a more comprehensive understanding of various domains.

\textbf{Reasoning:}
\begin{itemize}
\item \textit{Challenges in Strict Logical Deduction:} The models frequently encounter difficulties when attempting to perform strict logical deduction accurately. Common issues include ignoring premise conditions, misconstruing sufficient and necessary conditions, or making errors in logical chaining. 
These types of errors are commonly observed in manual analyses.

For instance, given a condition, ``\textit{If Myers is on the team, neither Ortega nor Paine can be}'', and a solution, ``\textit{Ortega, Paine, Thomson, and Zayre are on the team}'', the model incorrectly states that this solution is wrong because ``\textit{Paine and Ortega are on the team}'', neglecting to first satisfy the premise condition ``\textit{If Myers is on the team}''. Furthermore, the model demonstrates a misunderstanding of the difference between sufficient and necessary conditions in its explanation of another question and states: ``\textit{If Kayne is assigned to an ambassadorship, then so is Jaramillo. This constraint is essentially the same as the given constraint that if Jaramillo is assigned to one of the ambassadorships, then so is Kayne}''. 

To address these limitations, it is essential to improve the models' abilities to recognize and apply logical rules and refine their understanding of logical structures.

\item \textit{Difficulty with Counterfactual Reasoning:} The models consistently struggle with counterfactual reasoning tasks. They have difficulty generating alternative scenarios, evaluating hypothetical outcomes, or exploring potential consequences based on varying assumptions. For instance, the models frequently make incorrect judgments for counterfactual questions in the LSAT-AR task: ``\textit{Which one of the following, if substituted for the constraint that [Constraint A], would have the same effect in determining the assignment?}'' Enhancing the models' capabilities in handling counterfactual reasoning tasks is vital for developing a more comprehensive problem-solving skillset.

\item \textit{Struggles in Multi-hop Complex Reasoning:} The models have difficulty accurately executing multi-hop complex reasoning tasks, often displaying inconsistent logic, omitting inference steps, or producing flawed reasoning chains. To address a broader range of complex problems, it is crucial to improve the models' abilities to systematically navigate and process multi-step reasoning tasks.

\item \textit{Establishing Incorrect Conclusions and Contradictory Reasoning:} The models occasionally set an incorrect conclusion first and then generate contradictory reasoning based on that faulty foundation. This behavior emphasizes the need for improved reasoning verification and error correction techniques in the models' problem-solving processes.

\item \textit{Concealed Substitution of Concepts:} The models sometimes covertly substitute one concept with another similar one, leading to inaccurate or misleading reasoning. 
For example, in a biology exam, the model replaces the concept of ``\textit{isotopically labeled amino acids}'' with ``\textit{isotopically labeled tRNA (a tool for transporting amino acids)}'', resulting in erroneous reasoning.
This issue underscores the importance of better concept disambiguation and reasoning coherence in future model iterations. 

\item \textit{Difficulty in Identifying Solutions:} The models occasionally struggle to discover feasible solutions for specific problems, possibly due to limitations in their knowledge, reasoning capabilities, or problem-solving strategies. Addressing this shortcoming involves refining the models' ability to explore, evaluate, and select appropriate solutions based on the given problem context.

\item \textit{Vulnerability to Contextual Disturbance:} The reasoning ability of large language models is often easily disrupted by changes in the surrounding context. When the context is modified, the models may produce different deductions for the same condition, suggesting that the robustness of their reasoning ability is not yet sufficient. This observation emphasizes the need to develop models that can maintain consistent reasoning performance, even in the presence of varying contextual information, ensuring more reliable and stable problem-solving capabilities.
\end{itemize}

\textbf{Calculation:} The model is prone to making calculation errors, particularly when dealing with complex variable substitutions. This may be attributed to the inherent limitations of the model's computation process in handling mathematical operations, as well as its difficulty in parsing intricate relationships between variables. Consequently, the model may struggle to maintain accuracy and precision when attempting to solve problems involving advanced algebraic manipulations or multi-step calculations.
To address this issue, future iterations of the model should focus on enhancing its mathematical reasoning capabilities and improving its ability to recognize and apply relevant mathematical rules. This could involve incorporating specialized modules or mechanisms specifically designed to handle complex calculations, variable substitutions, and numerical problem-solving tasks. By refining the model's ability to accurately process and solve intricate mathematical problems, we can expand its applicability across a broader range of disciplines and domains, ensuring a more comprehensive and robust problem-solving skillset.

By addressing these reasoning weaknesses, future large language models can be developed with more robust problem-solving capabilities, enabling them to effectively tackle a broader range of human-centric tasks and exhibit more sophisticated reasoning skills that align closely with human cognition.

\section{Discussion about Future Directions}
In light of the findings and limitations identified in our analysis, we point out several potential future directions for the development of large foundation models. These directions aim to address the weaknesses observed and further improve the models' capabilities in various human-centric tasks.
\par{\textbf{Inclusion of External Knowledge and Formulas:}}
Enriching the models with external knowledge sources, like formulas and domain-specific knowledge can help enhance their performance in mathematical and knowledge-intensive tasks. 
Specifically, developing models that can effectively handle domain-specific tasks, such as those in law, biology, or physics, requires the integration of specialized knowledge bases and expertise into the model, and enables the model to adapt to different verticals more effectively.
This could involve integrating structured knowledge repositories, mathematical and scientific concepts  into the models with pre-training or knowledge-enhanced prompting methods, allowing them to access and apply relevant information more efficiently.

\par{\textbf{Strict Complex Logical Reasoning:}}
Improving the models' capacity for strict complex logical reasoning is crucial for their performance in a wide range of human-centric tasks. This could involve the creation of new datasets that emphasize complex reasoning, as well as incorporating APIs and external symbolic compilers that can execute strict logical or mathematical deduction, and use the execution results to further facilitate logical analysis and reasoning verification.

\par{\textbf{Multi-lingual Reasoning Capabilities Generalization:}}
As mentioned in Sec. \ref{sec:cot-analysis}, the reasoning capabilities of models are variant across different language, where the reasoning ability is relatively better for rich-resourced language like English. 
Enhancing the models' multi-lingual reasoning capabilities is essential for their applicability in a diverse range of real-world scenarios. 
Therefore, future directions can put more focus on enhancing multilingual generalization of reasoning capability of foundation models. 
\par{\textbf{Multi-modal Evaluation:}}
Expanding the evaluation framework to include multi-modal tasks can provide a more comprehensive assessment of the models' capabilities. This could involve incorporating visual, auditory, or interactive tasks that require the models to process and reason with multiple types of input simultaneously and generates multi-modal outputs for comprehensive real-world applications. In the future work, we will focus on the multi-modal version of \benchmarkname.
\par{\textbf{Better Automatic Evaluation Metrics for Human-centric Tasks:}}
Developing more robust and meaningful automatic evaluation metrics is crucial for the objective assessment of large language models' performance. Future research should focus on devising metrics that can accurately capture the models' understanding, knowledge, and reasoning abilities, while taking into account the nuances and complexities of real-world tasks.

\par{\textbf{Robustness of Reasoning Capability:}}
Improving the robustness of the models' reasoning capabilities is essential for ensuring their consistency and reliability across various contexts. This can be achieved by exploring techniques that enhance the models' ability to maintain consistent reasoning performance, even when faced with changes in the surrounding context or variations in the input data.

By addressing these future directions, foundation models can be further developed and refined to exhibit more advanced capabilities that align closely with human cognition, ultimately enabling them to tackle a broader range of complex, human-centric tasks with greater accuracy and reliability.

\section{Conclusion}

In this paper, we introduce \benchmarkname, a novel benchmark specifically designed to assess the general capabilities of large foundation models with respect to human-level cognition. The benchmark comprises high-quality official admission tests, qualification exams, and advanced competitions tailored for human participants, including law school admission tests and college entrance examinations.
These assessments establish officially recognized standards for gauging human capabilities, making them well-suited for evaluating foundation models in the context of human-centric tasks. Additionally, \benchmarkname~incorporates bilingual tasks in both Chinese and English, offering a more comprehensive assessment of model behavior. We have carried out an extensive evaluation of three cutting-edge large foundation models: Text-Davinci-003, ChatGPT, and GPT-4, using \benchmarkname. Remarkably, GPT-4 surpasses average human performance on LSAT, SAT, and math competition, attaining a 95\% accuracy rate on the SAT Math test and a 92.5\% accuracy on the Gaokao English test, demonstrating the impressive performance of contemporary foundation models.
Despite their significant achievements, our in-depth manual analyses also reveal the limitations of these large language models in terms of understanding, knowledge utilization, reasoning and calculation. Guided by these findings, we explore potential future research avenues in this domain. By assessing these foundation models on human-centric tasks and probing their capabilities more deeply, we strive to foster the development of models that are more closely aligned with human cognition. Ultimately, this will enable them to tackle a broader range of intricate, human-centric tasks with increased accuracy and reliability.

\bibliography{anthology,test}
\bibliographystyle{nips}


\appendix
\section{Data Examples}
\begin{figure*}[h]
    \centering
    \includegraphics[width=\textwidth]{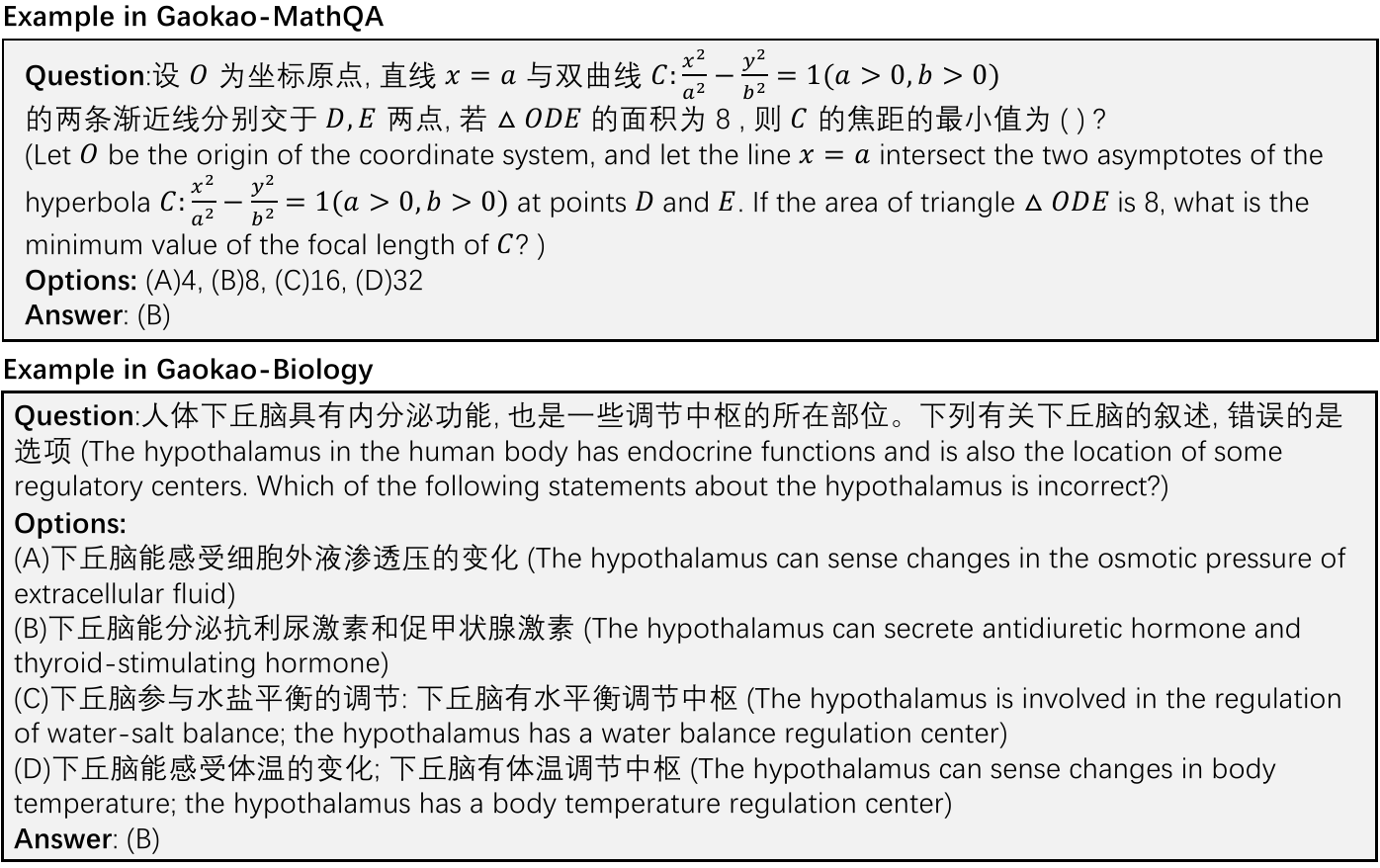}
    \caption{Data examples in Gaokao.}
    \label{fig:data-example}
\end{figure*}

\begin{figure*}[h]
    \centering
    \includegraphics[width=\textwidth]{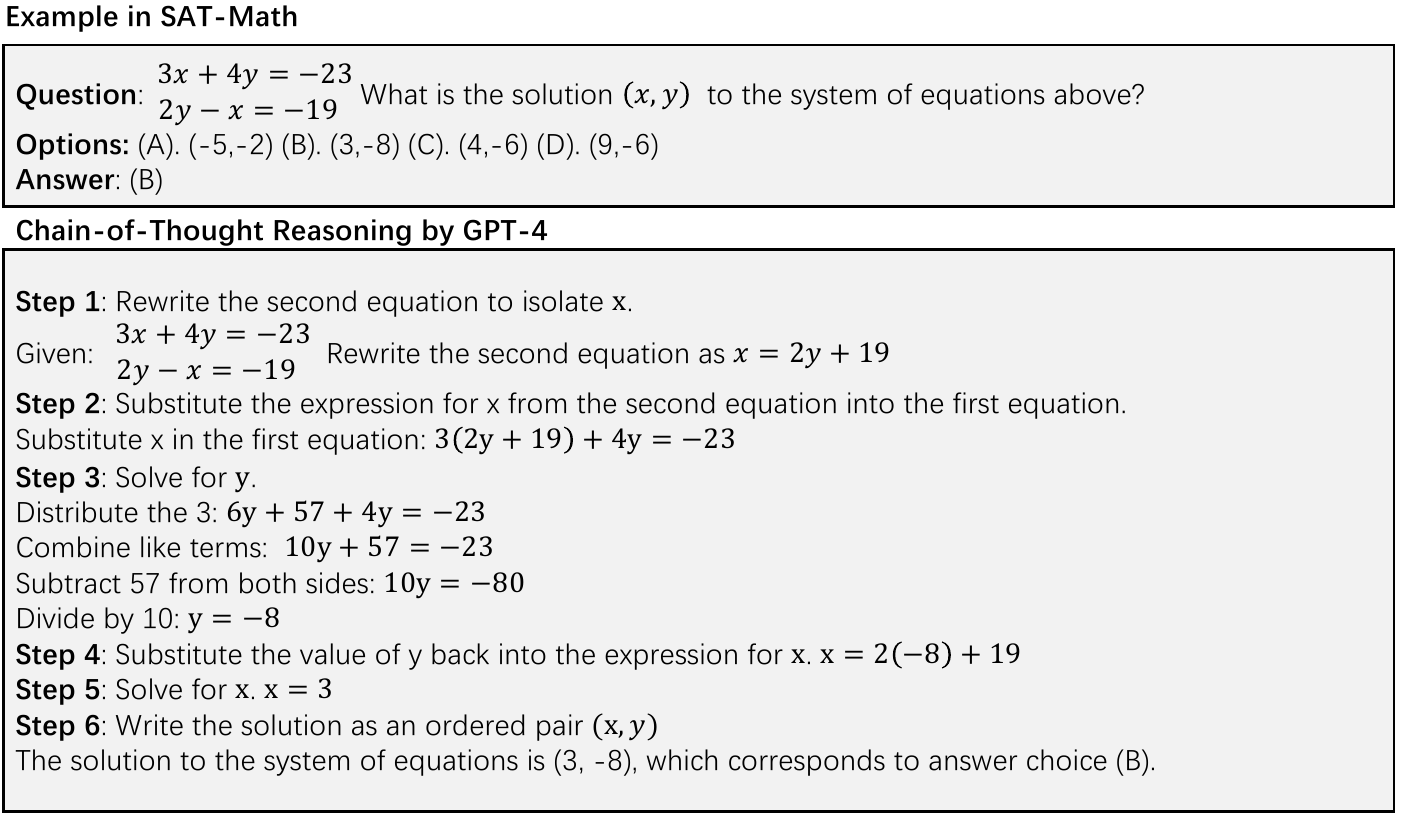}
    \caption{Data example in SAT and corresponding CoT reasoning process.}
    \label{fig:data-example-2}
\end{figure*}

Few data examples in Gaokao is shown in Fig. \ref{fig:data-example}, and an example in SAT and corresponding Chain-of-Thought reasoning process generated by GPT-4 is shown in Fig. \ref{fig:data-example-2}.
\section{Data Contamination Issue}
The issues surrounding data contamination and future web scrapes on training data for LLMs are noteworthy. Most of current benchmarks and datasets up to date suffer from these vulnerabilities.
To exam the situation of contamination, we provided timestamp for the 9 new Gaokao datasets and we can evaluate on the latest tests (later than 2022) released later than the training data timestamp of ChatGPT and GPT-4. 
Hereinafter, from AGIEval, we provide results comparing the GPT-4 zero-shot performance on six Gaokao subjects with and without risk of data contamination (Chinese, English, and History have not been included in this analysis due to the constrained size of the exams for these subjects). The uncontaminated dataset comprises entries released in 2022, which postdates the GPT-4 training data's timestamp (September 2021). The results are reported on Tab. \ref{tab:data-contamination}.
Evidently, we observe that barring the Mathematics subjects, the performance experiences a minor drop in the absence of contamination, yet remains proximate to the performances on the complete datasets. This finding substantiates that while AGIEval may similarly encounter data contamination, it still retains its value as a useful and effective human-centric benchmark for evaluating the abilities of foundation models against complex human-oriented tasks. 
\begin{table}[h]
\caption{Analysis on data contamination risk on AGIEval. The uncontaminated set includes examples released later than the time stamp of training data of ChatGPT and GPT-4.}
\label{tab:data-contamination}
\begin{tabular}{l|cccc}
\hline
                 & full-set-size & accuracy & uncontaminated-set-size & accuracy \\ \hline
Gaokao-geography & 199           & 76.9\%   & 37                      & 73\%     \\
Gaokao-biology   & 210           & 75.7\%   & 58                      & 77.6\%   \\
Gaokao-chemistry & 207           & 51.7\%   & 64                      & 42.2\%   \\
Gaokao-physics   & 200           & 39\%     & 5                       & 40\%     \\
Gaokao-MathQA    & 351           & 47\%     & 47                      & 29.8\%   \\
Gaokao-mathcloze & 118           & 16.1\%   & 25                      & 4\%      \\ \hline
\end{tabular}
\end{table}


\end{document}